\documentclass[11pt]{article}
\usepackage{listings}
\usepackage[preprint]{acl}

\usepackage{times}
\usepackage{latexsym}
\usepackage{amsmath}
\usepackage{amssymb}
\usepackage[T1]{fontenc}

\usepackage[utf8]{inputenc}

\usepackage{microtype}

\usepackage{inconsolata}

\usepackage{graphicx}

\title{Reasoning with Context Management and External Memory}

\author{Man-Ling Sung, William Hartmann, Jiangying Zhou
 \\
 RTX BBN Technologies \\ Cambridge MA, USA
 }

\begin{document}

\maketitle

\begin{abstract}


Reasoning models maintain information about the status of their reasoning process through in-context memory. Complex problems require larger context windows and memory requirements. In this scenario, context management without caching risks discarding useful information for reasoning. 
We propose RCMEM, an approach that extends context management with an external memory to maintain additional information without burdening the reasoning model with additional context. RCMEM significantly reduces computation time on problems that require repeated computation of intermediate steps, while achieving better performance on complex problems.
We demonstrate our approach on a few mathematics problems using model sizes ranging from 0.4 to 25 million parameters, and GPT-2.
\end{abstract}

\section{Introduction}
\label{intro}

Current work in reasoning models arose from \emph{Chain-of-Thought} (CoT) style approaches \cite{wei2022chain, zhang2024chain}.
By iteratively appending generated output to the context, a reasoning model can work through the steps to solve the problem.
A limitation of this approach is that the required memory grows linearly with the computation.
Longer context windows also come with attention limitations in terms of both efficiency \cite{keles2023computational} and accuracy \cite{liu2023lost}.
More sophisticated approaches use an agentic paradigm where information is passed between different models and tools \cite{lei2026memcot}.
Even with agentic approaches, it still remains a challenge to determine how much information to retain in the contextual memory of any one model.

Context memory management has recently become a topic of interest for LLMs \cite{yan2025inftythink, yang2025pencil, lu2025scaling}.
As an LLM reasons by adding additional information to the context window, it will eventually run out of memory.
At that point information begins to be removed from the context based on age, that can cause important information, e.g., instructions, to no longer be available and lead to catastrophic failures in the reasoning process.
Approaches to contextual memory management address the problem by proactively compress, summarize, and remove information in the context before memory issues arise.

A major limitation in contextual memory management alone is that the information disappears after summarization. 
This presents two key challenges. 
The first is in the case of a mistake or contradiction. If the model reaches a faulty conclusion, it is impossible to backtrack its reasoning process to discover and recover from the issue.
The second case is related to repeated subproblems. 
In order to solve some complex problems, it is required to solve subproblems multiple times (e.g., the recursive solution to the Fibonacci sequence). The worst-case time complexity without caching those intermediate solutions could increase exponentially.
To overcome these problems, recent works extend contextual memory management with external memory, providing theoretical proofs and applications to reinforcement learning. 

We specifically address the challenge of repeated computation of subproblems by extending contextual memory management with an external memory to supervised training. 
Starting from the PENCIL proposed by \citet{yang2025pencil} that applies reduction rules on the reasoning traces, we introduce special tokens that allow the model to read and write to an external memory. Our contributions are as follows: (1) We introduce RCMEM that integrates context management and external memory with reduction rules. (2) We show that the computation required for a model with external memory scales significantly slower for problems that require repeated computation.
(3) We verify our model's performance experimentally on several mathematical problems on supervised and fine-tuned models.

\section{Related Work}

\paragraph{Contextual memory management}
Standard CoT approaches \cite{wei2022chain} manage contextual memory by appending to the context.
Improvements to attention computation like Flash Attention \cite{dao2022flashattention}, or approximations like Linformer \cite{wang2020linformer} and Longformer \cite{beltagy2020longformer} allow for longer contexts without significantly increasing computation per token.
However, there are still issues with memory and model confusion \cite{liu2023lost}.
Contextual memory management compresses the information in the context to prevent memory issues.
Inftythink \cite{yan2025inftythink} uses instruction fine-tuning to adapt an LLM to iteratively summarize the existing context during the reasoning process.
PENCIL \cite{yang2025pencil} introduces special tokens and trains the LLM end-to-end to summarize intermediate reasoning steps.
SUPO \cite{lu2025scaling} also trains a model to summarize intermediate reasoning steps, but uses an reinforcement learning (RL) approach.
The summarization is part of the objective function and does not require training data with contextual memory summaries.

\paragraph{External memory}
An alternative approach is to offload contextual memory externally. For instance, retrieval-augmented generation (RAG,  \citet{guu2020retrieval}) gives the LLM access to information far beyond what can be stored in context, though selection of what to include in the context window is still a challenge \cite{barnett2024seven}.
For reasoning models, there is the additional challenge of deciding what information to offload to external memory.
Early work simply wrote information to an external scratchpad \cite{nye2021show}.
Since then, memory has been found to be critical to the performance of agentic systems \cite{zhang2025survey}.
Recent approaches have added native memory lookup beyond contextual memory, such as fine-tuning an LLM to produce memory access commands \cite{modarressi2023ret, modarressi2024memllm}, replaced feed-forward layers with locality-sensitive hashing \cite{zeng2023lookupffn}. 
Modern commercial LLMs have multiple layers of memory with agentic workflows to manage the flow of information \cite{lei2026memcot}. 
Engram \cite{cheng2026conditional} uses a modern n-gram lookup for phrase-level caching.
MeKi \cite{ding2026meki} uses a static memory at the token level to inject knowledge into the hidden layers of the network.


\paragraph{LLM reasoning with context compression and external memory}

Recent attempts in combining context management with external memory has been made. Neural Paging \cite{chen2026neural} provides a formal definition of the external memory concept in LLMs and offered theoretical analysis of complexity. Multiple possible realizations are tested, for example, Unified Memory Agent (UMA, \citet{zhang2026learning}) uses a dual architecture of compact core summary plus a structured Memory Bank with CRUD (create, read, update, and delete) operations trained end-to-end with RL. Dep-Search \cite{liu2026dep} combines dependency-aware reasoning traces with persistent memory using GRPO (Group Relative Policy Optimization) training. 



\section{Proposed methodology}

Our implementation is built on the reduction mechanism in PENCIL architecture, which offers a way to purge redundant information in the intermediate reasoning steps in CoT supervised training. This reduction mechanism complements the memory eviction in external memory by other reinforcement learning methods---clearing out expired reasoning traces that could lead to context burden in future computation. The combination of the two mechanisms, reduction of intermediate information in reasoning, and access to external memory, are the main originality of our approach.

\begin{figure}   \centering
  \includegraphics[width=0.35\paperwidth]{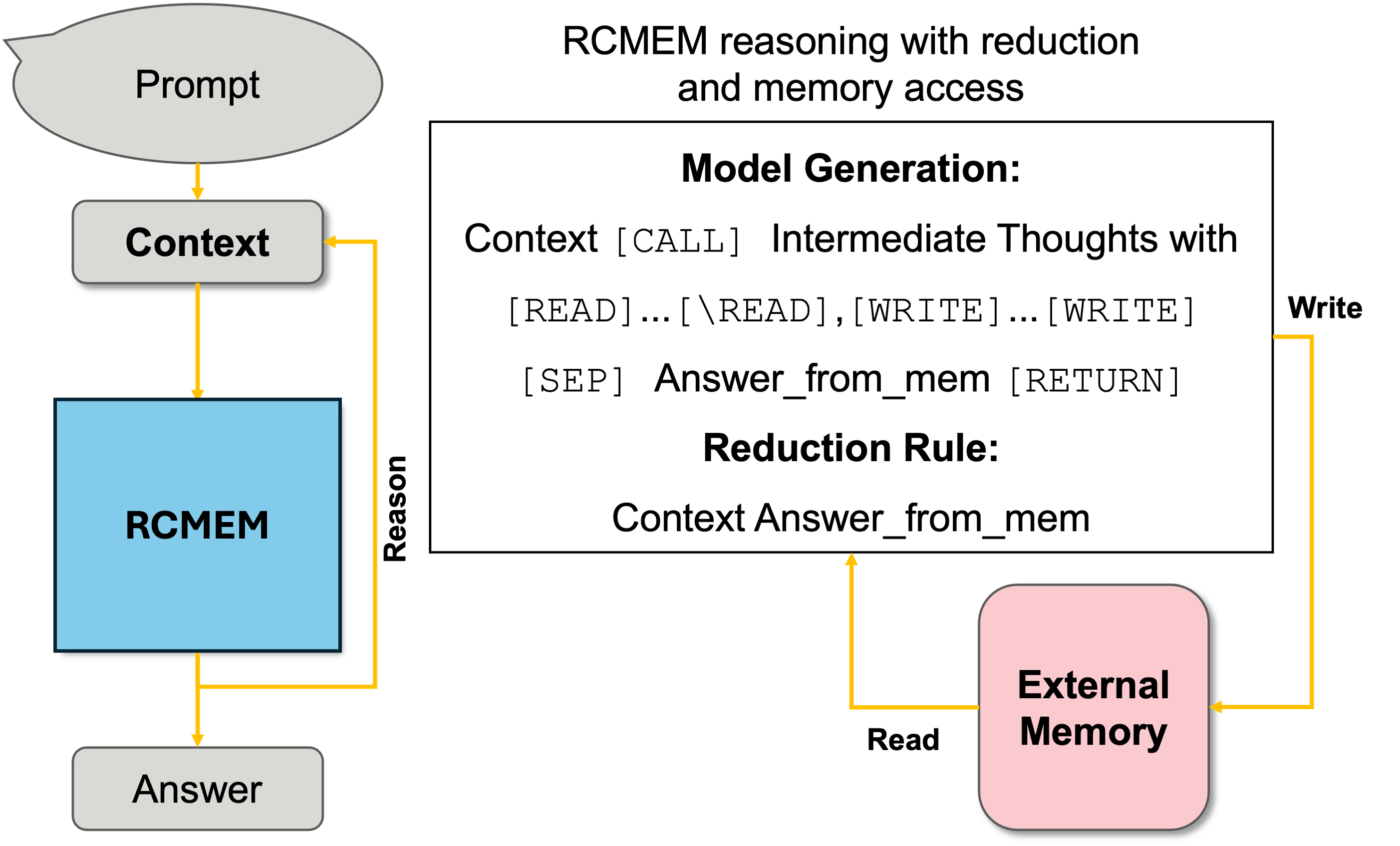}
  \caption{Illustration of how memory access is introduced on top of reduction rule in RCMEM}
  \label{fig:PENCIL_RW_design_rw}
\end{figure}

\subsection{Context management - PENCIL}

The original PENCIL \cite{yang2025pencil} approach introduces a reduction mechanism to the LLM autoregressive generation process, where intermediate thoughts are summarized recursively during inference. 
The reduction mechanism is achieved by adding three new tokens: \texttt{[CALL]}, \texttt{[SEP]} and \texttt{[RETURN]} to the CoT process, which separate the CoT sequence into context ($C$), thoughts ($T$), and answer ($A$). The process can be thought of as simulating a call stack, with the reduction rule illustrated as follows:
\begin{equation}
C\texttt{ [CALL] } T \texttt{ [SEP] } A \texttt{ [RETURN] } => C\, A
\end{equation}
Once a computation is completed (marked by \texttt{[RETURN]}), the reduction rule removes all intermediate reasoning steps $T$ (between \texttt{[CALL]} and \texttt{[SEP]}) and keeps only $A$ (between \texttt{[SEP]} and \texttt{[RETURN]}) as the remaining answer. This allows the model to reason recursively, and enables the LLM to perform more complex and longer reasoning with limited context length. It outperforms the standard CoT approach in complex reasoning problems like Einstein’s puzzles.

\subsection{External memory - RCMEM}

We extend this work by introducing an external memory with read and write access. In the remainder of the work we refer to the extension as RCMEM. Four additional tokens are introduced: \texttt{[READ]}, \texttt{[/READ]}, \texttt{[WRITE]}, and \texttt{[/WRITE]}. The model is trained to generate the additional tokens just as the original PENCIL model. The reduction rule operates identically to the PENCIL model. 

Given a memory $m$ with index $i$ and value $v$, we read a value  from the memory by calling an additional computation.
\begin{multline}
\texttt{[CALL] [READ] } m[i] \texttt{ [/READ]} \\
\texttt{[SEP] } v \texttt{ [RETURN]}
\end{multline}
The value $v$ between \texttt{[SEP]} and \texttt{[RETURN]} is recalled with the value stored in memory, and is kept for future computation after reduction. We write a new value to the memory by calling:
\begin{equation}
\texttt{[WRITE] } m[i] = v \texttt{ [/WRITE]} 
\end{equation}
Whenever the \texttt{[/READ]} or \texttt{[/WRITE]} tokens are generated, our memory access operation is invoked and the context is modified accordingly. Both memory access operations exist within the intermediate reasoning steps $T$, which are removed after the computation. 

\subsection{Computational improvements from external memory}
\label{computation}

The PENCIL model itself ``can perform universal space-efficient computation for solving any task \cite{yang2025pencil}.''
The introduction of an external memory does not fundamentally change the class of problems the model can solve.
What it provides is a more efficient approach to compute solution when problems are structured in a manner where repeated computation is required.

More formally, assume we have a problem that requires a search through every vertex in a directed acyclic graph (DAG).
If every vertex only has a single parent, then  the PENCIL approach can perform the optimal search through the graph (which would be a tree).
If, instead, the DAG is more general and each vertex can have multiple parents, then a vertex can be visited multiple times.
As the PENCIL approach provides no history of computation, every path through the DAG will be visited. Our extension with an external memory gives the model the ability to save the solutions to sub-graphs within the DAG.

Define a DAG as $G = (V,E)$, where $V$ is the set of vertices and $E$ is the set of edges.
Our model can evaluate $G$ in time polynomial in $|V| + |E|$ with an external memory of size $|V|$.
This is because every edge and every vertex must be visited once and the result of each visit to a vertex will be stored in memory.
The original PENCIL approach requires exponential time with respect to the size of the critical path in $G$ in the worst case.

Connecting to the knapsack problem in our experiments in Section \ref{sec:exp} , we can think of the solution as a DAG.
Each vertex is an entry in the DP matrix that represents the solution for the first $n$ items with a total weight of $w$.
Each edge is any valid connection between those entries, i.e., what other vertices are required in order to know the solution to the given subproblem.
The root node of the graph is the final solution using all $N$ items and the total weight $W$.

\subsection{Invalid memory access at inference}

One constraint in the design of our memory access is that the read and write operations require a specific syntax.
During inference, especially in early stages of training, it is possible for the model to generate output that is syntactically incorrect or to attempt to read from a memory location that does not exist.
We handle those scenarios as follows.
If the syntax is incorrect, then the memory access is skipped.
If the model attempts to read from null content, we remove the read operation and allow the model to continue processing by generating the solution. 
For details of these design decisions, please refer to Appendix \ref{sec:invalid_memory_access}. 

\section{Experiments}
\label{sec:exp}
\subsection{Problems Selection}
To demonstrate the capabilities of our model, we explore three problems - the knapsack problem, longest common subsequence (LCS) and graph shortest path (GSP). We train the model to solve both knapsack and LCS using a recursive approach, while GSP is solved with dynamic programming (DP). Each time the answer to a subproblem is discovered, the result is saved to an external memory.
When the solution to that subproblem is required, it is read from memory instead of being recomputed.

The pseudocode for generating the training traces are presented in Appendix \ref{sec:pseudocode}. Note that both LCS and GSP are more straightforward problems, where neighboring values are accessed in every iteration. The purpose is to understand the model's behaviors without running exhaustive experiments. The main experimental focus will be on the knapsack problem where the complexity of the problem and constraints require the model to learn effective access to the memory under specific conditions.


\textbf{Knapsack Problem: }\label{Knapsack_problem}
We consider specifically the 0-1 version with integer weights and capacity; This kind of problem requires solving repeated subproblems where RCMEM is designed to handle. 
We have a knapsack with a maximum capacity in terms of weight ($W$).
We also have a set of $N$ items, each with a value and a weight.
The goal is to find the optimal set of items that fit within the knapsack and maximize the value.
An exhaustive search is exponential in the number of items, with time complexity of $O(2^N)$.
We train the models with $N \in [4, 5, .., 10]$ items. For a sample with \(N\) items, item weights and prices are randomly selected integers from \([1, N]\), while the knapsack capacity \(W\) is randomly chosen from \([6, 30]\).

\textbf{Longest Common Subsequence (LCS): }
This is a simpler problem, where given two equal length sequences randomly constructed of `A', `C', `G', `T', the goal is to find the length of their longest common subsequence. A subsequence is a string generated from the original string by removing zero or more characters without changing the relative order of the remaining characters. For simplicity, the length $L$ is the only variable in this experiment. Both sequence lengths must be the same. The time complexity of the recursive solution is $O(2^{2L})$. We trained the model on length $L = {4,5,6,7}$.

\textbf{Graph Shortest Path (GSP):}   Additionally, to understand if RCMEM brings any benefit over PENCIL on dynamic programming (DP) solutions, we also experimented using the Floyd Warshall Algorithm in finding the shortest path in a graph. 
We compute the shortest paths between all pairs of vertices in a weighted graph. The graph is a 2D matrix of size $V \times V$. The algorithm exhaustively looks for intermediate nodes within the vertices that return the smallest total weight. The time complexity is $O(V^3)$ with a space complexity of $O(V^2)$. We trained the model on graphs with a number of vertices $V = {4,5,6}$, and graph weights randomly initialized from \([-1, 6]\).

\subsection{Training Data}


For each problem instance, we generate a solution trace by applying the iterative process described in Section \ref{Knapsack_problem}. From these traces, we create training samples consisting of the problem definition, constraints, the CoT reasoning, and the full solution trace. 
For PENCIL, we formulate the reasoning process as solving a chain of equations recursively. The equation is stored as an intermediate thought, while the answer is kept after the computation. For example, in Knapsack: 
\begin{equation}
\texttt{[CALL] } P(n, w) = \hat{v} \texttt{ [SEP] } \hat{v} \texttt{ [RETURN]}
\end{equation}
where \(n \in [1, N]\), \(w \in [0, W]\), and \(\hat{v}\) is the optimal value for the set \((n, w)\). 
For RCMEM, an additional read operation is inserted when an equation’s value is already available in memory, and a write operation is used when a calculated answer is ready to store. All other parts of the reasoning trace are identical to PENCIL. The pseudocode for generating the training traces are in Appendix \ref{sec:pseudocode}.

\subsection{Analysis on the problem complexity}
\label{sec:analysis_data_complexity}

To evaluate the solution complexity between the original PENCIL approach and our proposed modification, we measured the number of tokens in each solution. Identical problem sets are generated for the PENCIL model, but with no external memory access. Some of the reasoning process is repeated for recursive solutions. 

\begin{figure}[!htbp]  \centering
  \includegraphics[width=0.9\columnwidth]{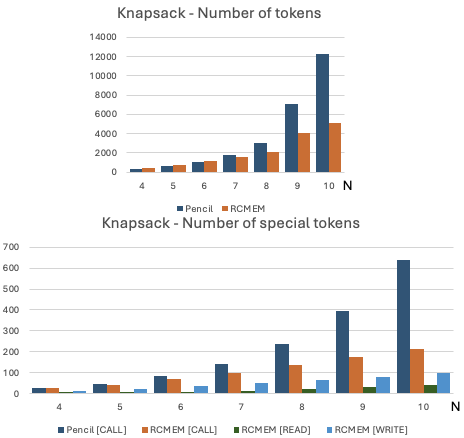}
    \caption{ (a) Average number of tokens and (b) special tokens for the Knapsack problem training sets. Note the number of tokens and \texttt{[CALL]} (iterations) of PENCIL exceed RCMEM at $N=7$ and $N=6$ respectively.}
   \label{fig:num_of_token_knapsack}
\end{figure}

\begin{figure}[!htbp]  \centering
  \includegraphics[width=\columnwidth]{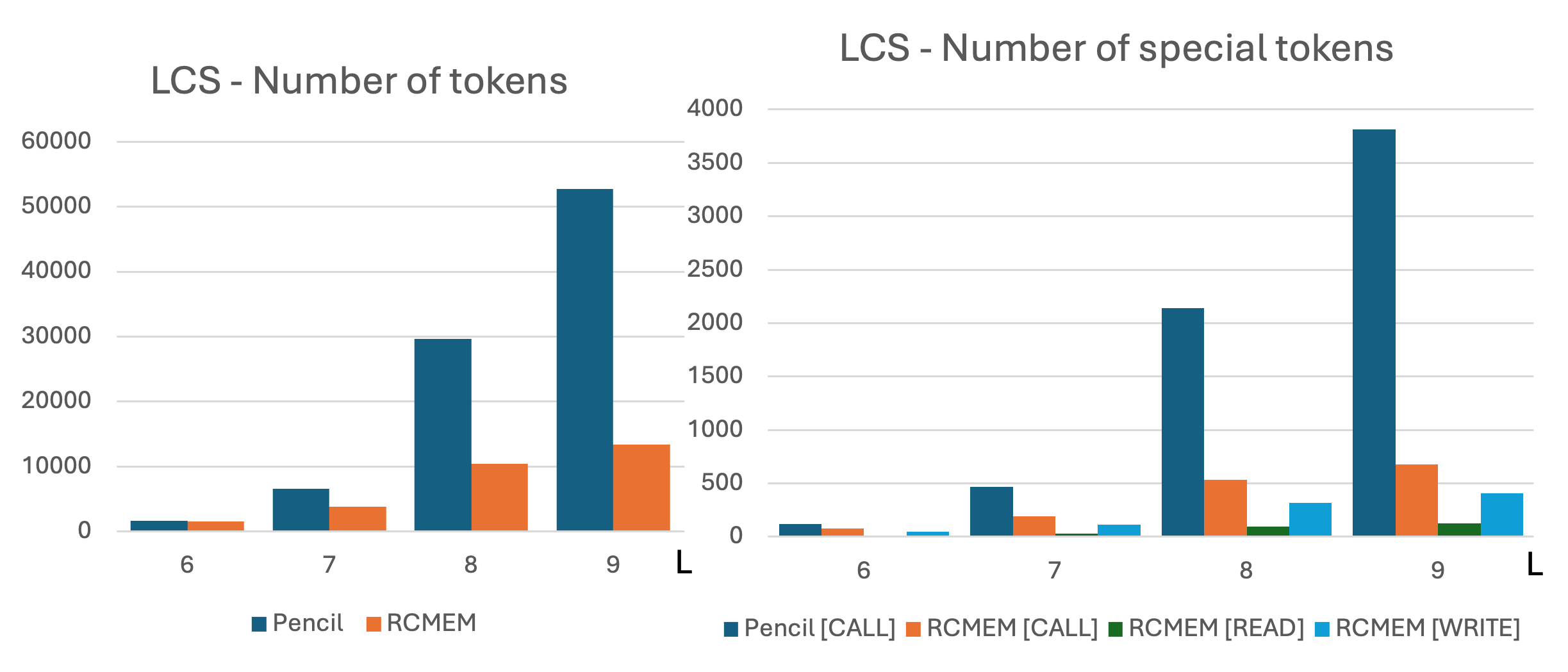}
  \caption{(a) Average number of tokens and (b) special tokens for LCS problem training sets. The number of tokens and iterations are much larger on PENCIL.}
  \includegraphics[width=0.9\columnwidth]{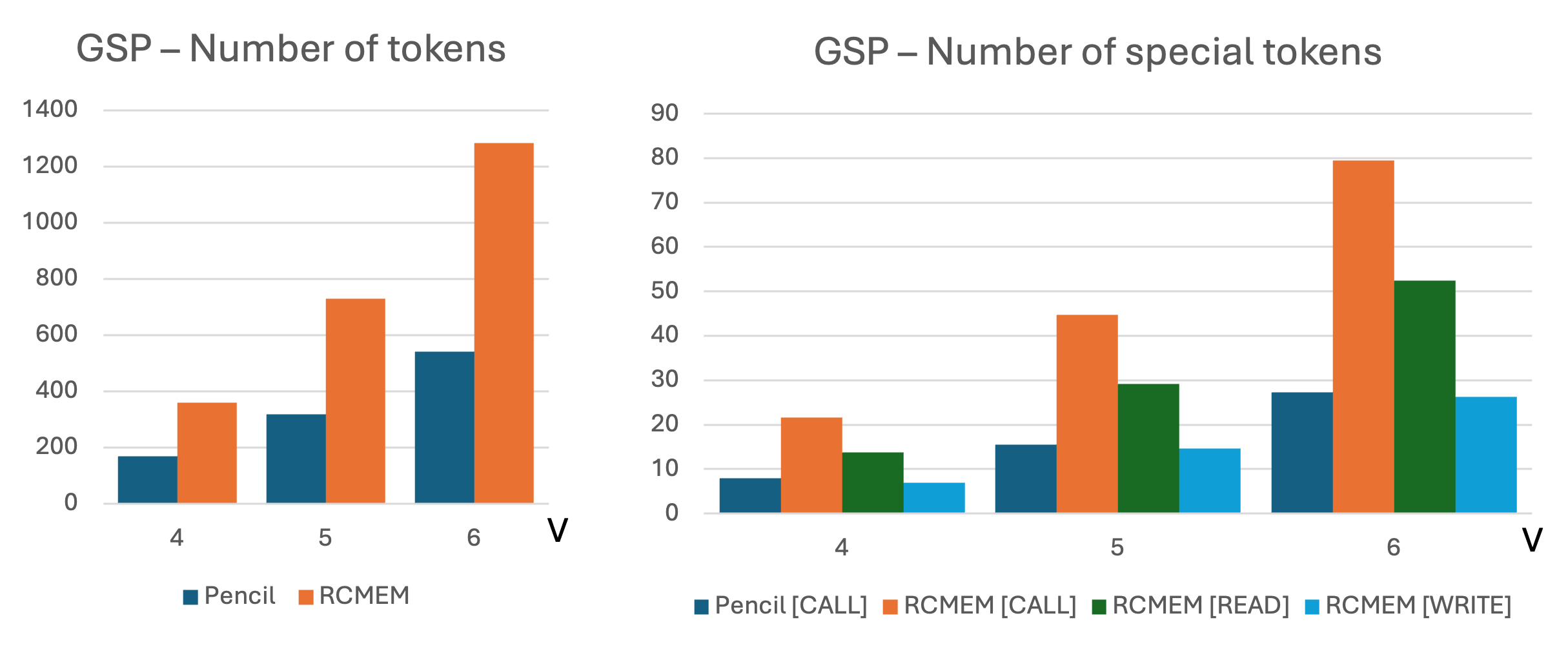}
  
  \caption{ (a) Average number of tokens and (b) special tokens for GSP problem training sets. The number of tokens and iterations are much larger on RCMEM.}
   \label{fig:num_of_token}
\end{figure}

\textbf{Knapsack: }
In theory the time complexity of recursion without memory access is $O(2^N)$, while having memory storage reduces the time complexity to $O(NW)$. 
Figure \ref{fig:num_of_token}a shows the average number of tokens in the training set. We observed that for smaller problem sizes the number of tokens from RCMEM is slightly larger than PENCIL until reaching a problem size of 7 items. After 7 items, the problem size grows exponentially as the number of items increases.
We also measure the number of \texttt{[CALL]}, \texttt{[READ]} and \texttt{[WRITE]} tokens to evaluate the effect of memory access in solving the problem in Figure \ref{fig:num_of_token}b. The number of \texttt{[CALL]} tokens measures the number of iterative CoT operations and the number of \texttt{[READ], [WRITE]} tokens measures the number of memory access operations. We observed that as the problem size grows larger, there is a significant reduction in CoT operations with increasing memory access in RCMEM. This shows that  the inherent complexity of the problem solution grows significantly slower when using external memory for intermediate computation.

\textbf{Longest Common Subsequence: }
Similar to knapsack, in theory the time complexity of the recursive approach is $O(2^{2L})$, while having memory storage reduces the time complexity to $O((2L)^2)$. 
As the problem size grows larger, the number of tokens and iterations grows exponentially. In all problem sizes generated, the number of tokens and iterations of RCMEM is consistently at least an order of magnitude smaller than PENCIL.

\textbf{Graph Shortest Path: }
Since the Floyd Warshall algorithm is a DP solution that reads the weights of the connecting vertices and calculates the updated weight of the target vertex, memory access does not simplify the reasoning process. Instead, RCMEM takes an extra two read operations (and iterations) to access the stored value, and one write operation to write into the memory. This tripled the number of iterations, and doubled the number of tokens for RCMEM compared to PENCIL. The only benefit of RCMEM over PENCIL is the ability to access values in the memory that are calculated earlier, instead of relying solely on the model's reasoning process. It evaluates whether memory itself, without the advantage of bypassing recursive computation, can still bring a benefit to the reasoning process.

\subsection{Model Training}

As with PENCIL, we adopt the learning setting where models train until convergence with unconstrained data access. This mirrors the common scenarios in language model training where data can be effectively infinite \cite{hoffmann2022training}. The sequence is truncated to the maximal context window that fits the model if it exceeds the model’s capacity. All hyperparameter settings are the same in all experiments, with the learning rate of 1e-3 decaying to 1e-4 in 20k steps with adamw optimizer. Depending on the problem and model size, the model runs on 1-2 A6000 GPUs. For each problem size, we generate 5 copies of the training, validation, and test set to train the models. The results are then averaged over the 5 runs. The model checkpoint that gives the highest accuracy on the validation set is used to score the test set. In addition to our proposed model, we also trained the PENCIL models with the same set of architectures. They are trained with PENCIL style data without access to the external memory. 

\subsubsection{Knapsack Problem}
\label{sec:knapsack_config}
For the knapsack problem, we explore various model sizes and training setups. The models are trained from scratch, and validated and tested on 100 samples each. We trained the models on $N=4$ to $10$, and experimented with various model sizes: 
 
\begin{itemize}
    \item 0.4M parameters model: Transformer with 2 layers, 2 heads, embedding size of 128.
    \item 3.17M parameters model: Transformer with 4 layers, 4 heads, embedding size of 256.
    \item 25.2M parameters model: Transformer with 8 layers, 8 heads, embedding size of 512.
\end{itemize}
For each model, we experimented with context windows of 256 and 512 tokens. 
We also experimented fine-tuning a pretrained GPT-2 model (124M, 2048 context length) using nanoGPT \cite{Karpathy2022nanogpt} with knapsack RCMEM data, to understand if the proposed method is applicable to LLM fine-tuning.

On a side note, CoT---which PENCIL and RCMEM are built on top of---is excluded from the analysis because it was already shown that PENCIL outperforms CoT in long reasoning tasks \cite{yang2025pencil}, which applies to the Knapsack problem. We also conducted a small experiment on CoT to validate this assumption in Appendix \ref{sec:cot_results}.

\subsubsection{Longest Common Subsequence}
For LCS, we only experiment on one model size. The model is trained from scratch and validated and tested on 100 samples each. We trained the model on sequences of $L=6$ to $9$ with each configuration. 
 
\begin{itemize}
    \item 0.4M parameters model: Transformer with 2 layers, 2 heads, embedding size of 128. Context length of 512.
\end{itemize}

\subsubsection{Graph Shortest Path}

The models are trained from scratch and validated and tested on 100 samples each. We trained the models on graph sizes of $V=4$ to $6$, and experimented with two model sizes: 
 
\begin{itemize}
    \item 3.17M parameters model: Transformer with 4 layers, 4 heads, embedding size of 256. Context length of 512.
    \item 25.2M parameters model: Transformer with 8 layers, 8 heads, embedding size of 512. Context length of 1024.
\end{itemize}

\subsection{Evaluation Metrics}

For all tasks, we measure the accuracy, which measures if the predicted answer matches ground truth. $y = \hat{y}$, with $y$ denoting the actual answer, and $\hat{y}$ the predicted answer. We also measure the average sample generation time
at inference to evaluate the efficiency of the models.

\section{Analysis}

We compared the performance of PENCIL and RCMEM with memory access on the Knapsack problem. The PENCIL approach uses the traditional recursive method to solve the Knapsack and the LCS problems, and the DP method on GSP. RCMEM creates an external memory that stores and updates the values. If the value is stored in memory, then the value is read from the memory instead of repeating the computation.

\subsection{Knapsack Problem}


\begin{figure}   \centering
  \includegraphics[width=0.35\paperwidth]{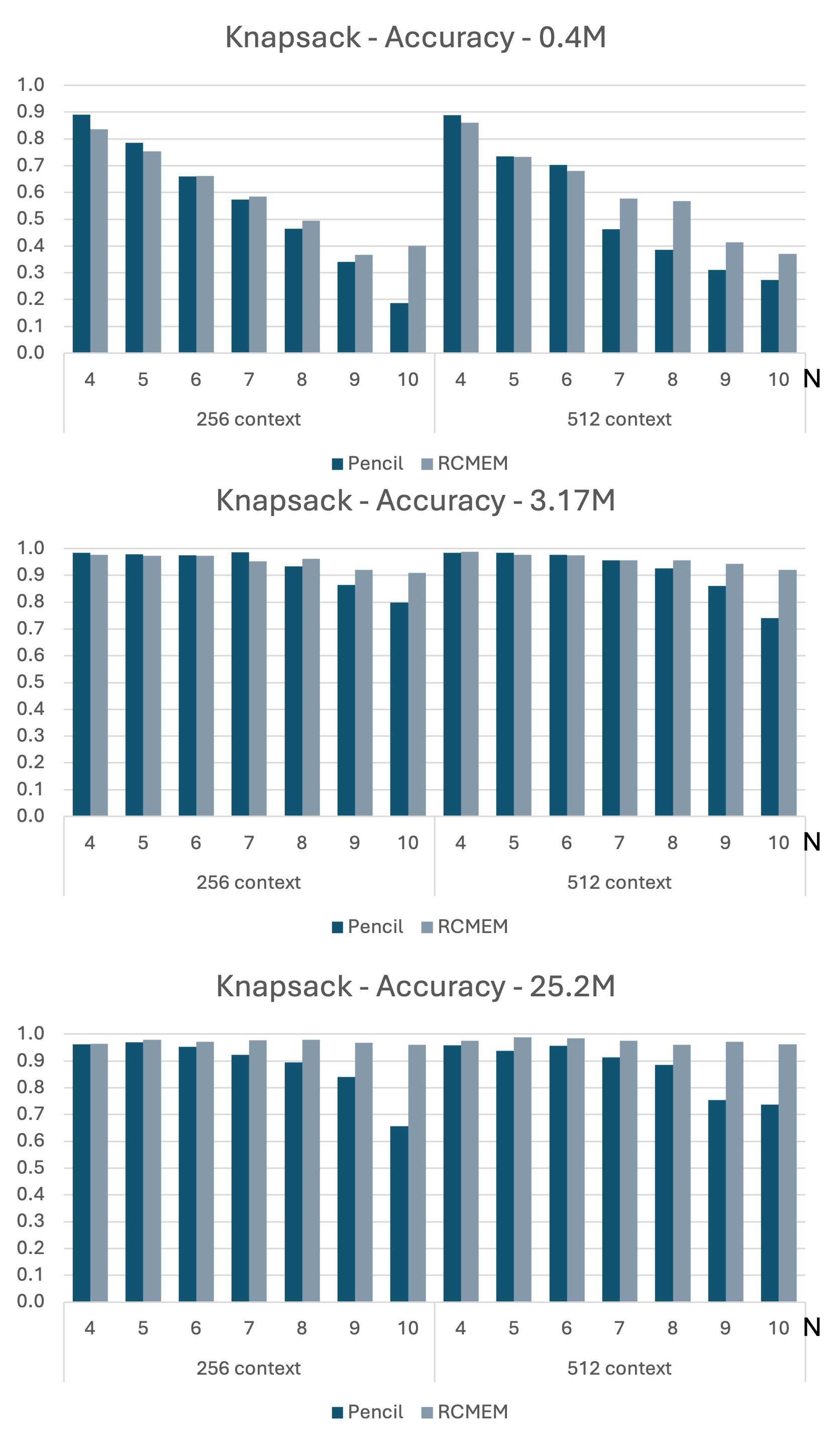}

  \caption{Accuracy of PENCIL and RCMEM on Knapsack across 3 model sizes. }
  \label{fig:Knapsack_accuracy}
  \end{figure}
  
\begin{figure} 
  \centering
  \includegraphics[width=0.35\paperwidth]{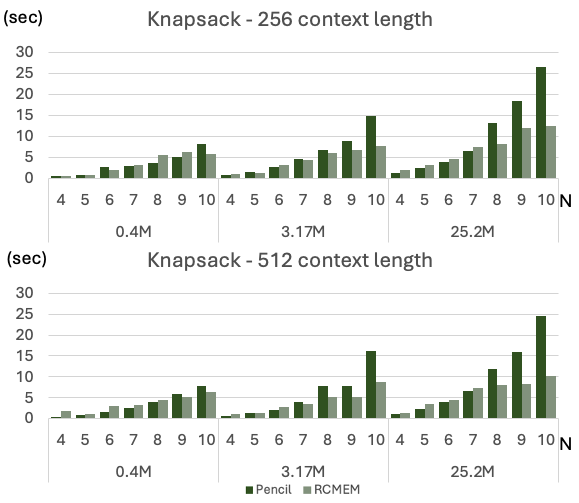}
  \caption{Average generation time for a sample at inference, on 256 and 512 context lengths.}
  \label{fig:Knapsack_time}
\end{figure}
\subsubsection{Accuracy}

Figure \ref{fig:Knapsack_accuracy} shows the average accuracy of PENCIL and RCMEM on the knapsack problem. 
Considering all models, the performance of the small model (0.4M) drops steadily as the problem size increases, with a ~40\% degradation across problem sizes of 4 to 10. For larger models (3.17M and 25.2M) the performance drop is mild---with $<5\%$ and $<10\%$ degradation, respectively. 
RCMEM outperforms PENCIL as problem complexity increases, regardless of context length. 
For the 3.17M model, RCMEM's advantage only becomes apparent with eight or more items, aligning with our complexity analysis in Section \ref{sec:analysis_data_complexity}. 
For the 25.2M model, RCMEM performance drops only marginally as complexity increases, whereas PENCIL shows a clear degradation. This suggests that larger models are better suited to learning additional memory access operations. 
Finally, the 0.4M model consistently maintains a 10-point lead from seven items onward with 512 context tokens. In resource-constrained environments (256 context tokens), RCMEM achieves double the accuracy of PENCIL at 10 items.
Our results suggests that for simpler problems, RCMEM gains no advantage from external memory access. However, for more complex problems, the benefits of memory access outweigh the increased operational complexity.

\subsubsection{Sample generation time} 

Figure \ref{fig:Knapsack_time} shows the average sample generation time at inference for models with context lengths of 256 and 512. The average generation time increases exponentially for PENCIL, while RCMEM increases at a lower rate with problem size. PENCIL outperforms RCMEM for simpler problems ($N<7$). At $N \geq 8$, the RCMEM generation time becomes much shorter (only half of PENCIL for 10 items).
This aligns with our analysis in Figure \ref{fig:num_of_token_knapsack} that the iterative process of RCMEM decreases exponentially compared to PENCIL after $N=7$, due to the benefit of external memory access. Fewer tokens means a faster response time of the model to an answer.

\subsubsection{Fine-tune GPT-2 model}

\begin{figure} 
  \centering
  \includegraphics[width=\columnwidth]{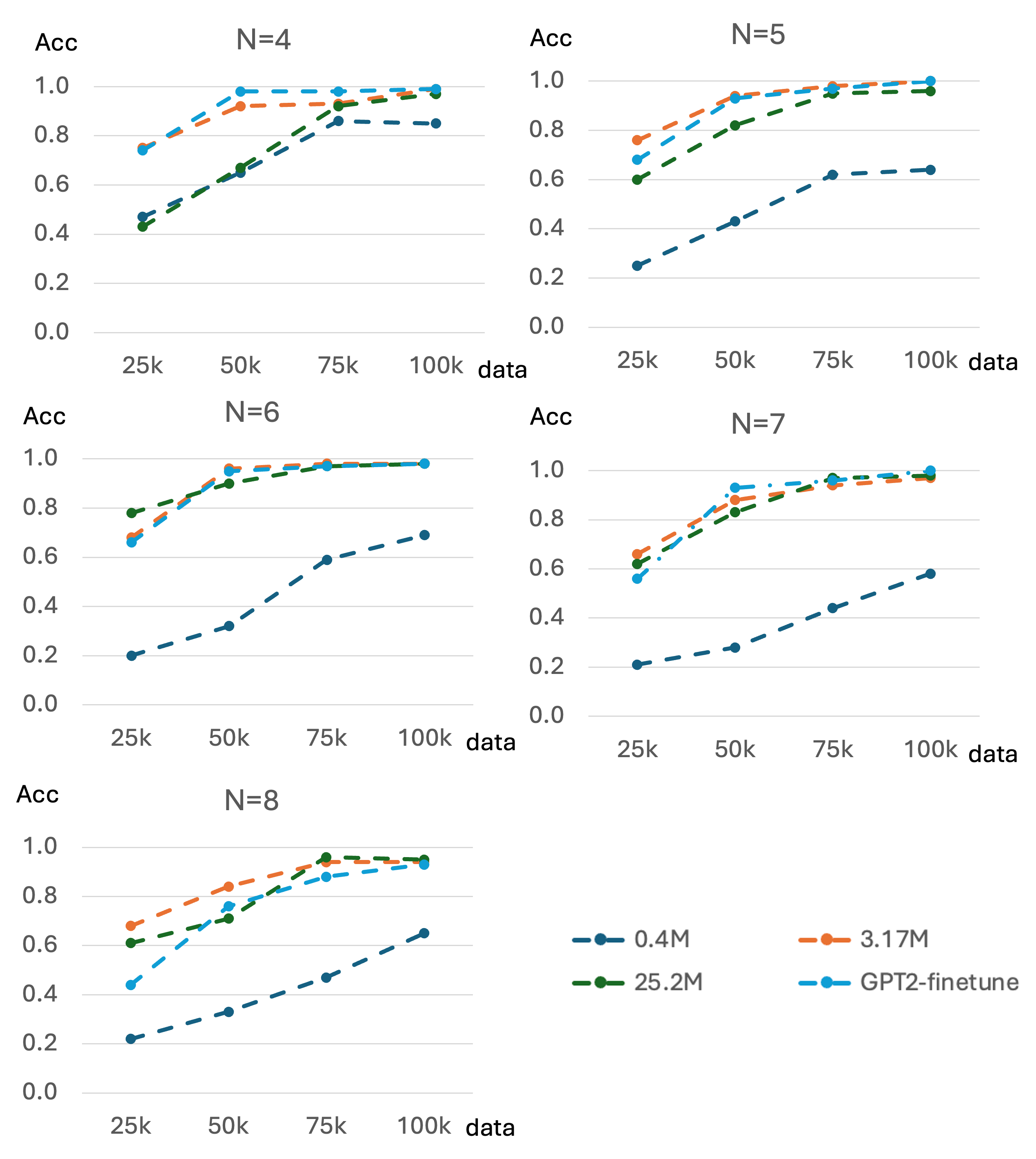}
  \caption{Knapsack - Accuracy of  0.4M, 3.17M, 25.2M and GPT-2 on 25k, 50k, 75k and 100k training samples.}
  \label{fig:GPT_comapre}
\end{figure}

We compare the convergence rate of one set of the seed dataset on the four models: 0.4M, 3.17M, 25.2M and GPT-2. The first three models are trained from scratch with the same architecture defined in Section \ref{sec:knapsack_config}, with 512 context length. The GPT-2 model is a pre-trained 1024M model with 2048 context length. The purpose of this experiment is to investigate the feasibility of applying RCMEM to fine-tune a pre-existing LLM that is trained on different data.

We compare the accuracy of the optimal models trained on 25k, 50k, 75k, and 100k training samples, presented in Figure \ref{fig:GPT_comapre}. Results show that although it takes more time to converge to the performance that other supervised models in the first 25k samples, GPT-2 is able to converge faster and outperform other smaller models with fine-tuning. For more complex problems ($N=8$), GPT-2 needs more training to outperform other models. This shows RCMEM is transferable to LLM and there is a benefit to fine-tuning from a pretrained LLM for faster convergence.

\subsection{Longest Common Subsequence}

\begin{figure}   \centering
\includegraphics[width=\columnwidth]{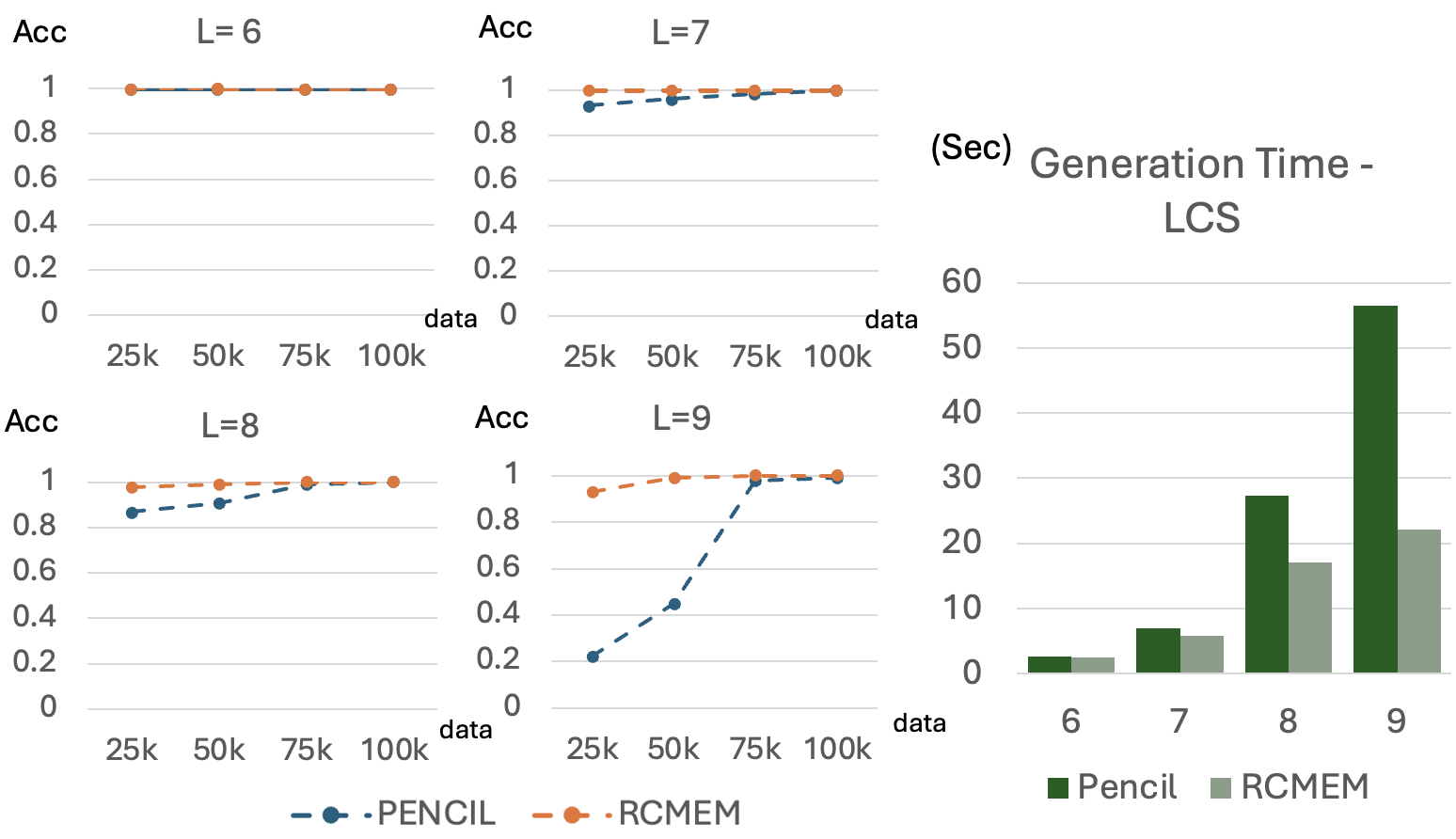}
\caption{Accuracy and sample generation time of PENCIL and RCMEM on LCS tasks. The accuracy are broken down to accuracy per batch of training samples to understand how fast the models are trained.}
\label{fig:accuracy_LCS}
\includegraphics[width=\columnwidth]{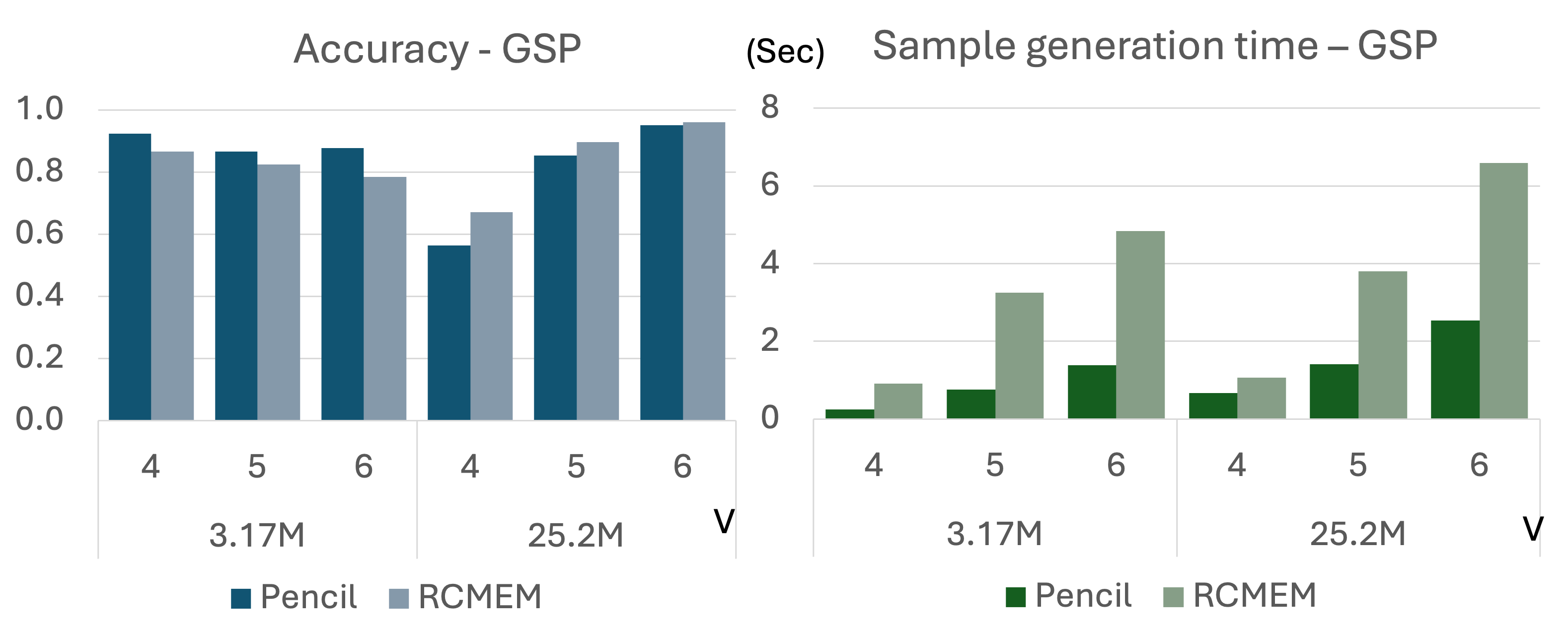}
\caption{Accuracy and sample generation time of PENCIL and RCMEM on GSP tasks. }
\label{fig:accuracy_GSP}
\end{figure}

\subsubsection{Accuracy}

Both PENCIL and RCMEM achieve 100\% accuracy after training on sufficient data on the 0.4M model. This shows that this is a much easier problem than Knapsack. So we analyze the amount of data they need to converge to an optimal solution. We present the accuracy of the models trained on 25k, 50k, 75k and 100k samples in figure \ref{fig:accuracy_LCS}. RCMEM coverages to 100\% accuracy much earlier than PENCIL. When $L$ increases, PENCIL needs more training samples to reach 100\% accuracy, while RCMEM can easily achieve and maintain such performance with only 25k or less training samples.

\subsubsection{Sample generation time} 
The sample generation time for PENCIL grows exponentially when the problem size increases. The time ratio difference with RCMEM matches with the number of tokens and iterations discussed in Figure \ref{fig:num_of_token}.
\subsection{Graph Shortest Path}

\subsubsection{Accuracy}
When the model size is small (3.17M with 256 context length), PENCIL consistently outperforms RCMEM. However, on a larger model (25.2M) with longer context length of 1024, RCMEM is able to outperform PENCIL on $V=4$, and has similar performance at $V=6$. We also notice the accuracy of the larger model is lower than that of a smaller model for $V=4$, but becomes better at $V=6$. This suggests that for a simpler problem, a large model might struggle from overfitting. In those scenarios, the ability to access the memory to correct a miscalculated answer from the model could help improve the performance. Also, even though the training data is more complex for RCMEM, it does not hurt the performance compared to PENCIL to use a smaller model when computation resource is limited.

\subsubsection{Sample generation time} 
The sample generation time for RCMEM is consistently roughly three times of PENCIL, regardless of the model size. This matches the ratio of the number of iterations discussed in Figure \ref{fig:num_of_token}.



\section{Conclusion}

We have introduced RCMEM, a model that extends the context management capabilities of PENCIL with an external memory for supervised training.
Our experiments have demonstrated that the external memory provides the ability for the model to remember previous computations without resorting to storing all intermediate steps in context.
On problems that can be represented as a directed acyclic graph, RCMEM requires a number of steps that is linear in the number of nodes and edges.
A reasoning model that relies on context management alone can require reasoning steps that are exponential in the length of the critical path of the graph.
Our experiments confirm that the time required grows slower for a model with external memory. The proposed methodology is transferable to a small LLM---GPT-2. Even on problems with dynamic programming solution, external memory can still benefit the reasoning process on large models.

\section{Limitations}

More work could be done to train the model to access external memory more effectively. In our experiments, we kept the setup comparable to PENCIL, with external‑memory access as the sole computational difference with PENCIL. 
With the current design, it is difficult for the model to learn when to access memory just by learning the pattern through token predictions. Future work on training specialized agentic AIs or applying the reduction rule to reinforcement learning to master the underlying procedural logic of memory access is needed.

As with PENCIL, RCMEM requires training data explicitly annotated with special tokens for effective memory access and resource management. While it is easy for deterministic algorithms like the Knapsack problem, developing automatic formation of reasoning processes is important in order to scale for complex, open-ended reasoning tasks. Future work will expand its use to real‑world applications where the reasoning steps are more complex---for example, resource allocation and logistics planning under computational constraints.

\bibliography{rscm.bib}
\newpage
\appendix
\section{Design decision of handling invalid memory access} \label{sec:invalid_memory_access}
We experimented various design in handling invalid memory access during inference of the generated tokens, as listed:
\begin{enumerate}

\item No memory access 
\begin{itemize}
\item This basically works as training the PENCIL model. For analysis purpose, the memory rate from PENCIL is inferred by extracting the intermediate solution at each trace stack on the generated text.
\end{itemize}
\item No invalid memory handle
\begin{itemize}
\item The read and write access are operated as it is. If invalid answer is read, the error propagates in the training.
\end{itemize}
\item Handle read only
\begin{itemize}
\item When the generated text tries to read from a non-existing memory, either if the weight exceeds the size of the memory, or the value of the item has not been written before (i.e. “\texttt{P(item, weight) = -1}” ). We reduce the text to exclude the \texttt{[SEP]} and \texttt{[RETURN]} tokens associate to the read operation to force the model to regenerate a valid answer. 
\end{itemize}
\item Bypass invalid READ memory access
\begin{itemize}
\item After the invalid memory access, the model stops trying to access the memory until reduction happens. The reason is to only access the memory again when the model might update the generated text and fix the issue after reduction.
\end{itemize}
\item Handle both read and write, Bypass invalid memory access
\begin{itemize}
\item In addition to (4), we implement the same handling to write operation. If the model tries to access from a non-existing memory, i.e. the weight exceeds the size of the memory. We reduce the text to exclude the \texttt{[SEP]} and \texttt{[RETURN]} tokens associate to the write operation. However, in theory it has less effect to the training process, due to the result of writing to the invalid memory is usually a chain effect of multiple wrong operations.
\end{itemize}
\end{enumerate}
We ran the setup on 0.4M model with 512 contextual tokens on knapsack, because of the observable degradation with $N = 4$ to $6$ makes the experiments more manageable. We trained and tested the models for 5 times and took the average on the accuracy. 

Besides accuracy, we also evaluate the performance of the models with trace rate and memory rate. With $x$ denotes the ground-truth reasoning trace, and $\hat{x}$ the model generated reasoning trace, $m$ denotes the ground truth memory and $\hat{m}$ the model induced memory. The metrics are calculated as follows:
\begin{itemize}
       \item Trace rate: Same metric used in \cite{yang2025pencil}, measures how the predicted tokens match with ground truth. $\frac{\sum_{i=1}^{min |x|,|\hat{x}|}{(x_{i} = {\hat{x}_{i}})}}{max|x|,|\hat{x}|}$ 
    \item Memory correct rate: Measures how the external memory, after read and write operations, matches with the ground truth. We only evaluate the memory that has been accessed.
    $\frac{\sum_{n=1, w=1}^{i\le N,j\le  W}{(m_{n,w}=\hat{m}_{n,w})}}{NW}$

\end{itemize}    

The average accuracy, memory correct rate and trace rate are presented in Table ~\ref{tab:memory_handle}. Actual memory correct rate is the accuracy of values in the memory actually access by the model. From the results, we observe the best accuracy and memory correct rate is obtained by just handling invalid read operations, with slight gain with early stopping on more items. Handling invalid write operations or not handling read operations hurt the performance in more complex problems. 


\begin{table*}[]
\centering 
\caption{Knapsack - Average accuracy and actual memory correct rate on 0.4M model with 512 content length, with different memory handle methods. R\&W - READ and WRITE, b.p. - bypass invalid memory access}
\label{tab:memory_handle}
\begin{tabular}{|c|c|c|c|c|c|}\hline
\textbf{\# of Item} & \textbf{No memory} & \textbf{Memory} & \textbf{Fix READ} & \textbf{Fix READ b.p.} & \textbf{Fix R\&W b.p. } \\ \hline
\multicolumn{6}{|c|}{\textbf{Accuracy}}  \\ \hline
4 & 0.816   & 0.74   & 0.86  & 0.824   & 0.832 \\ \hline
5 & 0.742   & 0.546  & 0.732 & 0.758   & 0.702 \\ \hline
6 & 0.584   & 0.284  & 0.68  & 0.7  & 0.55  \\ \hline
\multicolumn{6}{|c|}{\textbf{Memory Correct Rate}} \\ \hline
4 & 0.762   & 0.891  & 0.975 & 0.976   & 0.967 \\ \hline
5 & 0.570   & 0.901  & 0.954 & 0.958   & 0.842 \\ \hline
6 & 0.447   & 0.509  & 0.934 & 0.943   & 0.862 \\ \hline
\multicolumn{6}{|c|}{\textbf{Trace Rate}}    \\ \hline
4 & 0.781   & 0.789  & 0.752 & 0.770   & 0.785 \\ \hline
5 & 0.576   & 0.607  & 0.564 & 0.568   & 0.562 \\ \hline
6 & 0.444   & 0.441  & 0.430 & 0.405   & 0.421   \\\hline   
\end{tabular}
\end{table*}

\newpage

\section{Performance on Chain-of-Thought}

\label{sec:cot_results}

\begin{table}[]
\centering 
\caption{Knapsack - Accuracy of 3.17M on CoT, PENCIL and RCMEM, with 512 and 256 context tokens.}
\label{tab:cot_compare}
\begin{tabular}{|lccc|}
\hline
\multicolumn{4}{|c|}{\textbf{Model Size 3.17M}}                                                                                        \\ \hline
\multicolumn{1}{|l|}{\textbf{}}            & \multicolumn{3}{c|}{\textbf{Context 512}}                                                 \\ \hline
\multicolumn{1}{|l|}{\textbf{\# items}}    & \multicolumn{1}{c|}{\textbf{CoT}} & \multicolumn{1}{c|}{\textbf{PENCIL}} & \textbf{RCMEM} \\ \hline
\multicolumn{1}{|l|}{4}                    & \multicolumn{1}{c|}{0.776}        & \multicolumn{1}{c|}{0.984}           & 0.988          \\ \hline
\multicolumn{1}{|l|}{5}                    & \multicolumn{1}{c|}{0.354}        & \multicolumn{1}{c|}{0.984}           & 0.976          \\ \hline
\multicolumn{1}{|l|}{6}                    & \multicolumn{1}{c|}{0.222}        & \multicolumn{1}{c|}{0.976}           & 0.974          \\ \hline
\multicolumn{1}{|l|}{7}                    & \multicolumn{1}{c|}{0.086}        & \multicolumn{1}{c|}{0.956}           & 0.956          \\ \hline
\multicolumn{1}{|l|}{8}                    & \multicolumn{1}{c|}{0.056}        & \multicolumn{1}{c|}{0.926}           & 0.956          \\ \hline
\multicolumn{1}{|l|}{\textbf{}}            & \multicolumn{3}{c|}{\textbf{Context 256}}                                                 \\ \hline
\multicolumn{1}{|l|}{\textbf{\#    items}} & \multicolumn{1}{c|}{\textbf{CoT}} & \multicolumn{1}{c|}{\textbf{PENCIL}} & \textbf{RCMEM} \\ \hline
\multicolumn{1}{|l|}{4}                    & \multicolumn{1}{c|}{0.212}        & \multicolumn{1}{c|}{0.984}           & 0.976          \\ \hline
\multicolumn{1}{|l|}{5}                    & \multicolumn{1}{c|}{0.048}        & \multicolumn{1}{c|}{0.978}           & 0.972          \\ \hline
\multicolumn{1}{|l|}{6}                    & \multicolumn{1}{c|}{0.028}        & \multicolumn{1}{c|}{0.974}           & 0.972          \\ \hline
\multicolumn{1}{|l|}{7}                    & \multicolumn{1}{c|}{0.008}        & \multicolumn{1}{c|}{0.986}           & 0.952          \\ \hline
\multicolumn{1}{|l|}{8}                    & \multicolumn{1}{c|}{0.001}        & \multicolumn{1}{c|}{0.934}           & 0.963          \\ \hline
\end{tabular}
\end{table}

PENCIL significantly outperforming CoT has been discussed and proved on several reasoning tasks in \cite{yang2025pencil}. We believe it outperforms CoT in knapsack problem as well due to the nature of the problem. As a sanity check we also trained a 3.17M with 512 and 256 content length on CoT and compare with PENCIL and RCMEM. The training data of PENCIL is used to make sure the only different of CoT to PENCIL is the introduction of reduction rule.  Results presented in Table \ref{tab:cot_compare}.

\newpage

\section{Training PENCIL on RCMEM data}

\begin{table*}[] 
\centering
\caption{Knapsack - Accuracy, time (in seconds) and trace rate of PENCIL and RCMEM, trained on the same RCMEM data.}
\label{tab:PENCIL_train_on_rw}
\begin{tabular}{|c|ccc|ccc|}
\hline
\textbf{}      & \multicolumn{3}{c|}{\textbf{PENCIL}}                                                         & \multicolumn{3}{c|}{\textbf{RCMEM}}                                                      \\ \hline
\textbf{Items} & \multicolumn{1}{l|}{\textbf{Acc}} & \multicolumn{1}{l|}{\textbf{Time}} & \textbf{Trace Rate} & \multicolumn{1}{l|}{\textbf{Acc}} & \multicolumn{1}{l|}{\textbf{Time}} & \textbf{Trace Rate} \\ \hline
\textbf{4}     & \multicolumn{1}{l|}{0.984}        & \multicolumn{1}{l|}{1.124}         & 0.932               & \multicolumn{1}{l|}{0.988}        & \multicolumn{1}{l|}{0.896}         & 0.938               \\ \hline
\textbf{5}     & \multicolumn{1}{l|}{0.966}        & \multicolumn{1}{l|}{2.303}         & 0.839               & \multicolumn{1}{l|}{0.976}        & \multicolumn{1}{l|}{1.322}         & 0.833               \\ \hline
\textbf{6}     & \multicolumn{1}{l|}{0.938}        & \multicolumn{1}{l|}{2.267}         & 0.734               & \multicolumn{1}{l|}{0.974}        & \multicolumn{1}{l|}{2.438}         & 0.670               \\ \hline
\textbf{7}     & \multicolumn{1}{l|}{0.844}        & \multicolumn{1}{l|}{3.296}         & 0.587               & \multicolumn{1}{l|}{0.956}        & \multicolumn{1}{l|}{2.980}         & 0.578               \\ \hline
\textbf{8}     & \multicolumn{1}{l|}{0.794}        & \multicolumn{1}{l|}{4.238}         & 0.550               & \multicolumn{1}{l|}{0.956}        & \multicolumn{1}{l|}{5.986}         & 0.462               \\ \hline
\textbf{9}     & \multicolumn{1}{l|}{0.684}        & \multicolumn{1}{l|}{5.299}         & 0.483               & \multicolumn{1}{l|}{0.943}        & \multicolumn{1}{l|}{9.613}         & 0.410               \\ \hline
\textbf{10}    & \multicolumn{1}{l|}{0.638}        & \multicolumn{1}{l|}{5.848}         & 0.398               & \multicolumn{1}{l|}{0.918}        & \multicolumn{1}{l|}{6.972}         & 0.337               \\ \hline
\end{tabular}
\end{table*}

To verify the gain of the RCMEM is from the access to external memory, not primarily due to the shortened token size in the datasets, we trained the PENCIL model with the RCMEM datasets. We experiment on the 3.17M model with 512 contextual tokens, on knapsack with $N = 4$ to $10$.
Since the PENCIL approach does not have access to external memory, the read and write operations are not invoked.
Instead, the PENCIL model must predict the values that would have been read from memory.
This is possible because the versions of the model trained on the RCMEM data are trained to make those predictions; they are just not used when the model has access to external memory. 
We independently performed training and inference 5 times and present the average accuracy and generation time for a sample, presented in Table \ref{tab:PENCIL_train_on_rw}. The results show consistent gain in RCMEM over PENCIL, with the accuracy gap increasing as more items are introduced. This further proves the gain to the model with memory access. 

We also observe that the trace rate of the model with no memory access (the PENCIL method on the RCMEM datasets) is similar to other RCMEM variants, especially on small problems. The difference begins to increase on larger problems, where the example generation time of RCMEM also exceeds that of PENCIL. This is due to memory access operations in RCMEM that increases the computation time. 
This confirms that the drop in trace rate is not caused by memory access itself. A possible reason is that the introduction of READ and WRITE operations that increases the complexity of the learning process for the model.

\newpage

\section{PseudoCode of RCMEM Implementation of the problems}
\label{sec:pseudocode}
The pseudocodes of the RCMEM and PENCIL implementation of the 3 problems investigated in this work are presented in Table \ref{tab:memory_handle_kp}, \ref{tab:memory_handle_lcs}, \ref{tab:memory_handle_gsp}.
\begin{table*}[ht]
\centering 
\caption{Pseudocode of generating training traces for Knapsack problem}
\label{tab:memory_handle_kp}
\begin{tabular}{|p{15.5cm}|}
\hline
\begin{lstlisting}[basicstyle=\small]
def knapsackRec(item, W, msg, RCMEM=True):
    if (W > 0 and still_has_item):
        msg += " [CALL] P(item, W) = "
        if mem[item][W] exist: # directly read from memory
            msg = msg + "[CALL] [READ] P(item, W ) [/READ] "
            value = mem[item][W]
            msg = msg + ``[SEP] value [RETURN]"
            return value, msg 
        else:
            msg = msg + " P( item , W}) = "
            max_value = 0 
            if W < weight_of_item:
                msg = msg + "[CALL] "
                max_value, msg = knapsackRec(item_next, W, msg)
                # RCMEM only
                if RCMEM: 
                    msg += f" [WRITE] P( item, W ) = max_value [/WRITE]"
                    mem[item][W] = max_value
                msg += "[SEP] max_value [RETURN]"
            else:
                msg += "Max( [CALL]"
                value, msg = knapsackRec(item, W, msg)
                msg += "[CALL] "
                value2, msg = knapsackRec(item, W-weight_of_item, msg)
                max_value = np.max((value, value2 + value_of_item))
                msg += "+value_of_item"
 
                #RCMEM only
                if RCMEM: 
                    msg += " [WRITE] P( item , W ) = max_value [/WRITE]"
                    mem[item][W] = max_value
                msg += " [SEP] max_value [RETURN]"
 
            return max_value, msg
 
# Create problem set
item_names = N items
item_weights = random.choices([1..N], k=N)
item = {item_names, item_weights}
max_weight = random.randint(6, 30)

# Create memory table
mem = create_table(num_of_items, item_weights, max_weight)
 
# Start solving problem
msg = "[CALL]"
final_answer, final_msg = knapsackRec(item, max_weight, msg)
\end{lstlisting} 
\\\hline

\end{tabular}
\end{table*}

\begin{table*}[ht]
\centering 
\caption{Pseudocode of generating training traces for Longest Common Subsequence problem}
\label{tab:memory_handle_lcs}
\begin{tabular}{|p{15.5cm}|}
\hline
\begin{lstlisting}[basicstyle=\small]
def LCSRec(s1, s2, m, n, msg, RCMEM=True):
    if (m <=0) or (n <= 0):
        if mem[item][W] exist: # directly read from memory
            msg += "[CALL] P(m, n) [SEP] 0 [RETURN]" 
            return value, msg 
        msg = msg + " P(m,n) = "
        if s1[m-1] == s2[n-1]:
            msg += "P(m-1, n-1) + 1="
            # RCMEM only and value in memory
            if RCMEM and (mem[m-1][n-1] != -1):
                max_value = mem[m-1][n-1]
                msg += f"[CALL] [READ] P(m-1, n-1) [/READ]"
                msg += "[SEP] max_value [RETURN]"
            else:
                max_value, msg = lcsRec(s1, s2, m-1, n-1, msg)
                msg += "= max_value + 1"
            max_value += 1
            msg += "= max_value"
            #RCMEM only
            if RCMEM: 
                msg += "[WRITE] P(m,n) = max_value [/WRITE]"
                mem[m][n] = max_value
            msg += " [SEP] max_value [RETURN]"
        else:
       msg += "Max("
            value1, msg = LCSRec(s1, s2, m, n-1, msg)
            msg += " , "
            value2, msg = LCSRec(s1, s2, m-1, n, msg)
            max_value = np.max(value1, value2)
            msg += ") = max_value "
            #RCMEM only
            if RCMEM: 
                msg += " [WRITE] P(m,n) = max_value [/WRITE] "
                mem[m][n] = max_value
            msg += "[SEP] {max_value} [RETURN] "        
        return max_value, msg


# Create problem set
alphabet = ['A', 'C', 'G', 'T']
s1 = "".join(random.choice(alphabet) for _ in range(L))
s2 = "".join(random.choice(alphabet) for _ in range(L))

# Create memory table
mem = create_table(L)
 
# Start solving problem
msg = "[CALL]"
final_answer, final_msg = LCSRec(item, W, msg)
final_msg += f" [SEP] Final LCS is {final_answer} [RETURN]"

\end{lstlisting} 
\\\hline

\end{tabular}
\end{table*}

\begin{table*}[ht]
\centering 
\caption{Pseudocode of generating training traces for Graph Shortest Path problem}
\label{tab:memory_handle_gsp}
\begin{tabular}{|p{15.5cm}|}
\hline
\begin{lstlisting}[basicstyle=\small]
def FloydWarshall(graph, msg, RCMEM=True):
    V = len(graph)
    for k, i, j in zip(range(V), range(V), range(V)):
        if ((graph[i][j] == -1) and (graph[k][j] != -1 and graph[i][k] != -1)) \
           or ((graph[i][j] > (graph[i][k] + graph[k][j])) \
           and (graph[k][j] != -1 and graph[i][k] != -1)):
            msg += "[CALL] graph[i][j] = graph[i][k] + graph[k][j] = "
            if RCMEM:
                msg += f"[CALL] [READ] graph [{i}] [{k}] [/READ] "
                value = graph[i][k]
                msg += f"[SEP] value [RETURN] + "
                msg += f"[CALL] [READ] graph [{k}] [{j}] [/READ] "
                value = graph[k][j]
                msg += f"[SEP] value [RETURN] "
            graph[i][j] = graph[i][k] + graph[k][j]
            value = graph[i][j]
            if RCMEM: 
                msg += "[WRITE] graph[i][j] = value [/WRITE] "
            msg += "[SEP] value [RETURN] \n " 
 
    return graph, msg
 
# Create memory table
graph = np.random.randint(-1, 6, size=(V, V)) 
# Start solving problem
msg = "[CALL]"
final_answer, final_msg = FloydWarshall(graph, msg, RCMEM = True)
 

\end{lstlisting} 
\\\hline

\end{tabular}
\end{table*}

\section{Detail accuracy and sample generation time for all models and context lengths} \label{sec:detail_accuracy}

For the knapsack problem, we present the detailed mean and standard deviation accuracy of the PENCIL and RCMEM results in Table \ref{tab:acc_mean_std}, and the average sample generation time and its standard deviation for all models at all context lengths in Table \ref{tab:generate_time}.

\begin{table*}[ht]
\caption{Average accuracy and standard deviation of PENCIL and RCMEM, on 3 model sizes and 2 context lengths.}
\label{tab:acc_mean_std}
\centering
\begin{tabular}{|c|cccccccc|}
\hline
\textbf{} & \multicolumn{8}{c|}{\textbf{Model Size 25.2M}}                                                                                                                                                                                                                                                                             \\ \hline
          & \multicolumn{4}{c|}{\textbf{Context 512}}                                                                     & \multicolumn{4}{c|}{\textbf{Context 256}}                                                                     \\ \hline
          & \multicolumn{2}{c|}{\textbf{PENCIL}}                  & \multicolumn{2}{c|}{\textbf{RCMEM}}                   & \multicolumn{2}{c|}{\textbf{PENCIL}}                  & \multicolumn{2}{c|}{\textbf{RCMEM}}                   \\ \hline
N      & \multicolumn{1}{c|}{mean} & \multicolumn{1}{c|}{std}  & \multicolumn{1}{c|}{mean} & \multicolumn{1}{c|}{std}  & \multicolumn{1}{c|}{mean} & \multicolumn{1}{c|}{std}  & \multicolumn{1}{c|}{mean} & \multicolumn{1}{c|}{std}    \\ \hline
4         & \multicolumn{1}{c|}{0.96} & \multicolumn{1}{c|}{0.00} & \multicolumn{1}{c|}{0.98} & \multicolumn{1}{c|}{0.00} & \multicolumn{1}{c|}{0.96} & \multicolumn{1}{c|}{0.00} & \multicolumn{1}{c|}{0.96} & \multicolumn{1}{c|}{0.00}   \\ \hline
5         & \multicolumn{1}{c|}{0.94} & \multicolumn{1}{c|}{0.00} & \multicolumn{1}{c|}{0.99} & \multicolumn{1}{c|}{0.01} & \multicolumn{1}{c|}{0.97} & \multicolumn{1}{c|}{0.00} & \multicolumn{1}{c|}{0.98} & \multicolumn{1}{c|}{0.00} \\ \hline
6         & \multicolumn{1}{c|}{0.96} & \multicolumn{1}{c|}{0.01} & \multicolumn{1}{c|}{0.98} & \multicolumn{1}{c|}{0.00} & \multicolumn{1}{c|}{0.95} & \multicolumn{1}{c|}{0.01} & \multicolumn{1}{c|}{0.97} & \multicolumn{1}{c|}{0.00}  \\ \hline
7         & \multicolumn{1}{c|}{0.91} & \multicolumn{1}{c|}{0.00} & \multicolumn{1}{c|}{0.98} & \multicolumn{1}{c|}{0.00} & \multicolumn{1}{c|}{0.92} & \multicolumn{1}{c|}{0.00} & \multicolumn{1}{c|}{0.98} & \multicolumn{1}{c|}{0.00}   \\ \hline
8         & \multicolumn{1}{c|}{0.89} & \multicolumn{1}{c|}{0.01} & \multicolumn{1}{c|}{0.96} & \multicolumn{1}{c|}{0.00} & \multicolumn{1}{c|}{0.89} & \multicolumn{1}{c|}{0.01} & \multicolumn{1}{c|}{0.98} & \multicolumn{1}{c|}{0.00}  \\ \hline
9         & \multicolumn{1}{c|}{0.75} & \multicolumn{1}{c|}{0.02} & \multicolumn{1}{c|}{0.97} & \multicolumn{1}{c|}{0.00} & \multicolumn{1}{c|}{0.84} & \multicolumn{1}{c|}{0.02} & \multicolumn{1}{c|}{0.97} & \multicolumn{1}{c|}{0.01}   \\ \hline
10        & \multicolumn{1}{c|}{0.74} & \multicolumn{1}{c|}{0.01} & \multicolumn{1}{c|}{0.97} & \multicolumn{1}{c|}{0.00} & \multicolumn{1}{c|}{0.66} & \multicolumn{1}{c|}{0.01} & \multicolumn{1}{c|}{0.96} & \multicolumn{1}{c|}{0.01}   \\ \hline
          & \multicolumn{8}{c|}{\textbf{Model Size 3.17M}}                                                                                                                                                                                                                                                                             \\ \hline
          & \multicolumn{4}{c|}{\textbf{Context 512}}                                                                     & \multicolumn{4}{c|}{\textbf{Context 256}}                                                         \\ \hline
          & \multicolumn{2}{c|}{\textbf{PENCIL}}                  & \multicolumn{2}{c|}{\textbf{RCMEM}}                   & \multicolumn{2}{c|}{\textbf{PENCIL}}                  & \multicolumn{2}{c|}{\textbf{RCMEM}}  \\ \hline
N      & \multicolumn{1}{c|}{mean} & \multicolumn{1}{c|}{std}  & \multicolumn{1}{c|}{mean} & \multicolumn{1}{c|}{std}  & \multicolumn{1}{c|}{mean} & \multicolumn{1}{c|}{std}  & \multicolumn{1}{c|}{mean} & \multicolumn{1}{c|}{std}    \\ \hline
4         & \multicolumn{1}{c|}{0.98} & \multicolumn{1}{c|}{0.00} & \multicolumn{1}{c|}{0.99} & \multicolumn{1}{c|}{0.00} & \multicolumn{1}{c|}{0.98} & \multicolumn{1}{c|}{0.00} & \multicolumn{1}{c|}{0.98} & \multicolumn{1}{c|}{0.00}  \\ \hline
5         & \multicolumn{1}{c|}{0.98} & \multicolumn{1}{c|}{0.00} & \multicolumn{1}{c|}{0.98} & \multicolumn{1}{c|}{0.00} & \multicolumn{1}{c|}{0.98} & \multicolumn{1}{c|}{0.00} & \multicolumn{1}{c|}{0.97} & \multicolumn{1}{c|}{0.00}   \\ \hline
6         & \multicolumn{1}{c|}{0.98} & \multicolumn{1}{c|}{0.00} & \multicolumn{1}{c|}{0.97} & \multicolumn{1}{c|}{0.00} & \multicolumn{1}{c|}{0.97} & \multicolumn{1}{c|}{0.00} & \multicolumn{1}{c|}{0.97} & \multicolumn{1}{c|}{0.00}  \\ \hline
7         & \multicolumn{1}{c|}{0.96} & \multicolumn{1}{c|}{0.00} & \multicolumn{1}{c|}{0.96} & \multicolumn{1}{c|}{0.00} & \multicolumn{1}{c|}{0.99} & \multicolumn{1}{c|}{0.00} & \multicolumn{1}{c|}{0.95} & \multicolumn{1}{c|}{0.00}  \\ \hline
8         & \multicolumn{1}{c|}{0.93} & \multicolumn{1}{c|}{0.01} & \multicolumn{1}{c|}{0.96} & \multicolumn{1}{c|}{0.00} & \multicolumn{1}{c|}{0.93} & \multicolumn{1}{c|}{0.01} & \multicolumn{1}{c|}{0.96} & \multicolumn{1}{c|}{0.01}   \\ \hline
9         & \multicolumn{1}{c|}{0.86} & \multicolumn{1}{c|}{0.01} & \multicolumn{1}{c|}{0.94} & \multicolumn{1}{c|}{0.01} & \multicolumn{1}{c|}{0.86} & \multicolumn{1}{c|}{0.01} & \multicolumn{1}{c|}{0.92} & \multicolumn{1}{c|}{0.00}   \\ \hline
10        & \multicolumn{1}{c|}{0.74} & \multicolumn{1}{c|}{0.01} & \multicolumn{1}{c|}{0.92} & \multicolumn{1}{c|}{0.00} & \multicolumn{1}{c|}{0.80} & \multicolumn{1}{c|}{0.01} & \multicolumn{1}{c|}{0.91} & \multicolumn{1}{c|}{0.01}   \\ \hline
\textbf{} & \multicolumn{8}{c|}{\textbf{Model Size 0.4M}}                                                                                                                                                                                                                                                                              \\ \hline
\textbf{} & \multicolumn{4}{c|}{\textbf{Context 512}}                                                                     & \multicolumn{4}{c|}{\textbf{Context 256}}                                                  \\ \hline
\textbf{} & \multicolumn{2}{c|}{\textbf{PENCIL}}                  & \multicolumn{2}{c|}{\textbf{RCMEM}}                   & \multicolumn{2}{c|}{\textbf{PENCIL}}                  & \multicolumn{2}{c|}{\textbf{RCMEM}}                    \\ \hline
N      & \multicolumn{1}{c|}{mean} & \multicolumn{1}{c|}{std}  & \multicolumn{1}{c|}{mean} & \multicolumn{1}{c|}{std}  & \multicolumn{1}{c|}{mean} & \multicolumn{1}{c|}{std}  & \multicolumn{1}{c|}{mean} & \multicolumn{1}{c|}{std}   \\ \hline
4         & \multicolumn{1}{c|}{0.89} & \multicolumn{1}{c|}{0.01} & \multicolumn{1}{c|}{0.86} & \multicolumn{1}{c|}{0.01} & \multicolumn{1}{c|}{0.89} & \multicolumn{1}{c|}{0.01} & \multicolumn{1}{c|}{0.84} & \multicolumn{1}{c|}{0.01}  \\ \hline
5         & \multicolumn{1}{c|}{0.73} & \multicolumn{1}{c|}{0.02} & \multicolumn{1}{c|}{0.73} & \multicolumn{1}{c|}{0.01} & \multicolumn{1}{c|}{0.79} & \multicolumn{1}{c|}{0.01} & \multicolumn{1}{c|}{0.75} & \multicolumn{1}{c|}{0.01}   \\ \hline
6         & \multicolumn{1}{c|}{0.70} & \multicolumn{1}{c|}{0.02} & \multicolumn{1}{c|}{0.68} & \multicolumn{1}{c|}{0.02} & \multicolumn{1}{c|}{0.66} & \multicolumn{1}{c|}{0.01} & \multicolumn{1}{c|}{0.66} & \multicolumn{1}{c|}{0.01}  \\ \hline
7         & \multicolumn{1}{c|}{0.46} & \multicolumn{1}{c|}{0.05} & \multicolumn{1}{c|}{0.58} & \multicolumn{1}{c|}{0.04} & \multicolumn{1}{c|}{0.57} & \multicolumn{1}{c|}{0.07} & \multicolumn{1}{c|}{0.58} & \multicolumn{1}{c|}{0.01} \\ \hline
8         & \multicolumn{1}{c|}{0.39} & \multicolumn{1}{c|}{0.03} & \multicolumn{1}{c|}{0.57} & \multicolumn{1}{c|}{0.02} & \multicolumn{1}{c|}{0.46} & \multicolumn{1}{c|}{0.03} & \multicolumn{1}{c|}{0.50} & \multicolumn{1}{c|}{0.01}  \\ \hline
9         & \multicolumn{1}{c|}{0.31} & \multicolumn{1}{c|}{0.06} & \multicolumn{1}{c|}{0.41} & \multicolumn{1}{c|}{0.02} & \multicolumn{1}{c|}{0.34} & \multicolumn{1}{c|}{0.04} & \multicolumn{1}{c|}{0.36} & \multicolumn{1}{c|}{0.02}   \\ \hline
10        & \multicolumn{1}{c|}{0.28} & \multicolumn{1}{c|}{0.02} & \multicolumn{1}{c|}{0.38} & \multicolumn{1}{c|}{0.02} & \multicolumn{1}{c|}{0.19} & \multicolumn{1}{c|}{0.04} & \multicolumn{1}{c|}{0.40} & \multicolumn{1}{c|}{0.03}   \\ \hline
\end{tabular}
\end{table*}


\begin{table*}[ht]
\centering 
\caption{Average generation time and standard deviation (in seconds) for a sample at inference, on 3 model sizes and 2 context lengths.}
\label{tab:generate_time}

\begin{tabular}{|c|cccccccc|}
\hline
\textbf{} & \multicolumn{8}{c|}{\textbf{Model Size 25.2M}}                                                                                                                                                           \\ \hline
\textbf{} & \multicolumn{4}{c|}{\textbf{Context 512}}                                                                   & \multicolumn{4}{c|}{\textbf{Context 256}}                                                  \\ \hline
\textbf{} & \multicolumn{2}{c|}{\textbf{PENCIL}}                 & \multicolumn{2}{c|}{\textbf{RCMEM}}                  & \multicolumn{2}{c|}{\textbf{PENCIL}}                 & \multicolumn{2}{c|}{\textbf{RCMEM}} \\ \hline
N         & \multicolumn{1}{c|}{mean} & \multicolumn{1}{c|}{std} & \multicolumn{1}{c|}{mean} & \multicolumn{1}{c|}{std} & \multicolumn{1}{c|}{mean} & \multicolumn{1}{c|}{std} & \multicolumn{1}{c|}{mean}   & std   \\ \hline
4         & \multicolumn{1}{c|}{1.2}  & \multicolumn{1}{c|}{0.3} & \multicolumn{1}{c|}{1.5}  & \multicolumn{1}{c|}{0.5} & \multicolumn{1}{c|}{1.4}  & \multicolumn{1}{c|}{0.2} & \multicolumn{1}{c|}{1.9}    & 0.2   \\ \hline
5         & \multicolumn{1}{c|}{2.4}  & \multicolumn{1}{c|}{0.4} & \multicolumn{1}{c|}{3.5}  & \multicolumn{1}{c|}{0.1} & \multicolumn{1}{c|}{2.4}  & \multicolumn{1}{c|}{0.4} & \multicolumn{1}{c|}{3.1}    & 0.7   \\ \hline
6         & \multicolumn{1}{c|}{4.1}  & \multicolumn{1}{c|}{1.1} & \multicolumn{1}{c|}{4.4}  & \multicolumn{1}{c|}{0.8} & \multicolumn{1}{c|}{3.9}  & \multicolumn{1}{c|}{0.8} & \multicolumn{1}{c|}{4.6}    & 1.1   \\ \hline
7         & \multicolumn{1}{c|}{6.7}  & \multicolumn{1}{c|}{1.5} & \multicolumn{1}{c|}{7.4}  & \multicolumn{1}{c|}{0.9} & \multicolumn{1}{c|}{6.6}  & \multicolumn{1}{c|}{1.1} & \multicolumn{1}{c|}{7.5}    & 1.2   \\ \hline
8         & \multicolumn{1}{c|}{12}   & \multicolumn{1}{c|}{2.6} & \multicolumn{1}{c|}{8}    & \multicolumn{1}{c|}{2.2} & \multicolumn{1}{c|}{13.2} & \multicolumn{1}{c|}{0}   & \multicolumn{1}{c|}{8.3}    & 1.9   \\ \hline
9         & \multicolumn{1}{c|}{15.9} & \multicolumn{1}{c|}{3.7} & \multicolumn{1}{c|}{8.4}  & \multicolumn{1}{c|}{2}   & \multicolumn{1}{c|}{18.5} & \multicolumn{1}{c|}{4.9} & \multicolumn{1}{c|}{12.1}   & 1.8   \\ \hline
10        & \multicolumn{1}{c|}{24.6} & \multicolumn{1}{c|}{2.4} & \multicolumn{1}{c|}{10.2} & \multicolumn{1}{c|}{1.1} & \multicolumn{1}{c|}{26.7} & \multicolumn{1}{c|}{8.4} & \multicolumn{1}{c|}{12.5}   & 5.1   \\ \hline
\textbf{} & \multicolumn{8}{c|}{\textbf{Model Size 3.17M}}                                                                                                                                                           \\ \hline
\textbf{} & \multicolumn{4}{c|}{\textbf{Context 512}}                                                                   & \multicolumn{4}{c|}{\textbf{Context 256}}                                                  \\ \hline
\textbf{} & \multicolumn{2}{c|}{\textbf{PENCIL}}                 & \multicolumn{2}{c|}{\textbf{RCMEM}}                  & \multicolumn{2}{c|}{\textbf{PENCIL}}                 & \multicolumn{2}{c|}{\textbf{RCMEM}} \\ \hline
N         & \multicolumn{1}{c|}{mean} & \multicolumn{1}{c|}{std} & \multicolumn{1}{c|}{mean} & \multicolumn{1}{c|}{std} & \multicolumn{1}{c|}{mean} & \multicolumn{1}{c|}{std} & \multicolumn{1}{c|}{mean}   & std   \\ \hline
4         & \multicolumn{1}{c|}{0.7}  & \multicolumn{1}{c|}{0.1} & \multicolumn{1}{c|}{1.1}  & \multicolumn{1}{c|}{0.1} & \multicolumn{1}{c|}{0.9}  & \multicolumn{1}{c|}{0}   & \multicolumn{1}{c|}{1}      & 0.2   \\ \hline
5         & \multicolumn{1}{c|}{1.3}  & \multicolumn{1}{c|}{0.3} & \multicolumn{1}{c|}{1.4}  & \multicolumn{1}{c|}{0.3} & \multicolumn{1}{c|}{1.5}  & \multicolumn{1}{c|}{0.1} & \multicolumn{1}{c|}{1.4}    & 0.2   \\ \hline
6         & \multicolumn{1}{c|}{2.2}  & \multicolumn{1}{c|}{0.4} & \multicolumn{1}{c|}{2.8}  & \multicolumn{1}{c|}{0.3} & \multicolumn{1}{c|}{2.7}  & \multicolumn{1}{c|}{0.2} & \multicolumn{1}{c|}{3.1}    & 0.2   \\ \hline
7         & \multicolumn{1}{c|}{4}    & \multicolumn{1}{c|}{0.6} & \multicolumn{1}{c|}{3.5}  & \multicolumn{1}{c|}{0.8} & \multicolumn{1}{c|}{4.6}  & \multicolumn{1}{c|}{0.3} & \multicolumn{1}{c|}{4.4}    & 0.2   \\ \hline
8         & \multicolumn{1}{c|}{7.9}  & \multicolumn{1}{c|}{0.6} & \multicolumn{1}{c|}{5.1}  & \multicolumn{1}{c|}{1.1} & \multicolumn{1}{c|}{6.7}  & \multicolumn{1}{c|}{1.4} & \multicolumn{1}{c|}{6}      & 0.4   \\ \hline
9         & \multicolumn{1}{c|}{7.8}  & \multicolumn{1}{c|}{0.3} & \multicolumn{1}{c|}{5.1}  & \multicolumn{1}{c|}{4.1} & \multicolumn{1}{c|}{8.9}  & \multicolumn{1}{c|}{1.9} & \multicolumn{1}{c|}{6.8}    & 1.1   \\ \hline
10        & \multicolumn{1}{c|}{16.1} & \multicolumn{1}{c|}{3.7} & \multicolumn{1}{c|}{8.9}  & \multicolumn{1}{c|}{0.5} & \multicolumn{1}{c|}{14.8} & \multicolumn{1}{c|}{4.5} & \multicolumn{1}{c|}{7.7}    & 1.8   \\ \hline
\textbf{} & \multicolumn{8}{c|}{\textbf{Model Size 0.4M}}                                                                                                                                                            \\ \hline
\textbf{} & \multicolumn{4}{c|}{\textbf{Context 512}}                                                                   & \multicolumn{4}{c|}{\textbf{Context 256}}                                                  \\ \hline
\textbf{} & \multicolumn{2}{c|}{\textbf{PENCIL}}                 & \multicolumn{2}{c|}{\textbf{RCMEM}}                  & \multicolumn{2}{c|}{\textbf{PENCIL}}                 & \multicolumn{2}{c|}{\textbf{RCMEM}} \\ \hline
N         & \multicolumn{1}{c|}{mean} & \multicolumn{1}{c|}{std} & \multicolumn{1}{c|}{mean} & \multicolumn{1}{c|}{std} & \multicolumn{1}{c|}{mean} & \multicolumn{1}{c|}{std} & \multicolumn{1}{c|}{mean}   & std   \\ \hline
4         & \multicolumn{1}{c|}{0.5}  & \multicolumn{1}{c|}{0}   & \multicolumn{1}{c|}{1.9}  & \multicolumn{1}{c|}{2.6} & \multicolumn{1}{c|}{0.5}  & \multicolumn{1}{c|}{0.1} & \multicolumn{1}{c|}{0.5}    & 0.1   \\ \hline
5         & \multicolumn{1}{c|}{0.9}  & \multicolumn{1}{c|}{0.1} & \multicolumn{1}{c|}{1.2}  & \multicolumn{1}{c|}{0.1} & \multicolumn{1}{c|}{0.9}  & \multicolumn{1}{c|}{0.2} & \multicolumn{1}{c|}{0.9}    & 0.3   \\ \hline
6         & \multicolumn{1}{c|}{1.6}  & \multicolumn{1}{c|}{0.1} & \multicolumn{1}{c|}{3}    & \multicolumn{1}{c|}{2.2} & \multicolumn{1}{c|}{2.7}  & \multicolumn{1}{c|}{0.3} & \multicolumn{1}{c|}{1.8}    & 0.1   \\ \hline
7         & \multicolumn{1}{c|}{2.5}  & \multicolumn{1}{c|}{0.4} & \multicolumn{1}{c|}{3.3}  & \multicolumn{1}{c|}{0.3} & \multicolumn{1}{c|}{3}    & \multicolumn{1}{c|}{0.4} & \multicolumn{1}{c|}{3.3}    & 2.4   \\ \hline
8         & \multicolumn{1}{c|}{3.9}  & \multicolumn{1}{c|}{0.9} & \multicolumn{1}{c|}{4.6}  & \multicolumn{1}{c|}{2.6} & \multicolumn{1}{c|}{3.6}  & \multicolumn{1}{c|}{3}   & \multicolumn{1}{c|}{8.7}    & 2.1   \\ \hline
9         & \multicolumn{1}{c|}{5.8}  & \multicolumn{1}{c|}{2}   & \multicolumn{1}{c|}{5.1}  & \multicolumn{1}{c|}{2.3} & \multicolumn{1}{c|}{5.2}  & \multicolumn{1}{c|}{1.8} & \multicolumn{1}{c|}{6.4}    & 0.9   \\ \hline
10        & \multicolumn{1}{c|}{7.9}  & \multicolumn{1}{c|}{1.5} & \multicolumn{1}{c|}{6.5}  & \multicolumn{1}{c|}{2.6} & \multicolumn{1}{c|}{8.3}  & \multicolumn{1}{c|}{1.8} & \multicolumn{1}{c|}{5.9}    & 2     \\ \hline
\end{tabular}
\end{table*}




\end{document}